\documentclass[lettersize,journal]{IEEEtran}
\usepackage{amsmath,amsfonts}
\usepackage{algorithmic}
\usepackage{algorithm}
\usepackage{array}
\usepackage[caption=false,font=normalsize,labelfont=sf,textfont=sf]{subfig}
\usepackage{textcomp}
\usepackage{stfloats}
\usepackage{url}
\usepackage{verbatim}
\usepackage{graphicx}
\usepackage{cite}
\usepackage{booktabs}
\usepackage{makecell,multirow,diagbox}
\usepackage{amsmath}
\usepackage{amssymb}

\usepackage{colortbl} 
\usepackage[dvipsnames]{xcolor}

\usepackage{tabularx}

\newcommand{\ie}{\textit{i}.\textit{e}.}
\newcommand{\eg}{\textit{e}.\textit{g}.}

\usepackage{hyperref}
\hypersetup{
    colorlinks=true,
    linkcolor=blue,
    filecolor=blue,      
    urlcolor=blue,
    citecolor=cyan,
}

\begin{document}

\title{CRF360D: Monocular 360 Depth Estimation via Spherical Fully-Connected CRFs}

 \author{Zidong Cao, Lin Wang* 
 \thanks{*Corresponding author.}
  \thanks{Z. Cao is with HKUST(GZ), China}
 \thanks{L. Wang is with HKUST(GZ) and HKUST, China}
 }

\markboth{Journal of \LaTeX\ Class Files,~Vol.~14, No.~8, August~2021}%
{Shell \MakeLowercase{\textit{et al.}}: A Sample Article Using IEEEtran.cls for IEEE Journals}


\maketitle

\begin{abstract}
Monocular $360^{\circ}$ depth estimation is challenging due to the inherent distortion of the equirectangular projection (ERP). This distortion causes a problem: 
spherical adjacent points are separated after being projected to the ERP plane, particularly in the polar regions.
To tackle this problem, recent methods calculate the spherical neighbors in the tangent domain.
However, as the tangent patch and sphere only have one common point, these methods construct neighboring spherical relationships around the common point.
In this paper,
we propose spherical fully-connected CRFs (\textbf{SF-CRFs}). We begin by evenly partitioning an ERP image with regular windows, where windows at the equator involve broader spherical neighbors than those at the poles. To improve the spherical relationships, our SF-CRFs enjoy two key components. Firstly, to involve sufficient spherical neighbors, we propose a Spherical Window Transform (\textbf{SWT}) module. This module aims to replicate the equator window's spherical relationships to all other windows, leveraging the rotational invariance of the sphere. Remarkably, the transformation process is highly efficient, 
completing the transformation of all windows in a $512 \times 1024$ ERP with 0.038 seconds on CPU. Secondly, we propose a Planar-Spherical Interaction (\textbf{PSI}) module to facilitate the relationships between regular and transformed windows, which not only preserves the local details but also captures global structures.
By building a decoder based on the SF-CRFs blocks, we propose \textbf{CRF360D}, a novel $360^{\circ}$ depth estimation framework that achieves state-of-the-art performance across diverse datasets. Our CRF360D is compatible with different perspective image-trained backbones (\eg, EfficientNet), serving as the encoder. \textit{For the demo and appendix, please check the project page at \url{https://vlislab22.github.io/CRF360D/}}.
\end{abstract}


\section{Introduction}
\label{sec:intro}

\begin{figure}[t]
    \centering
\includegraphics[width=\linewidth]{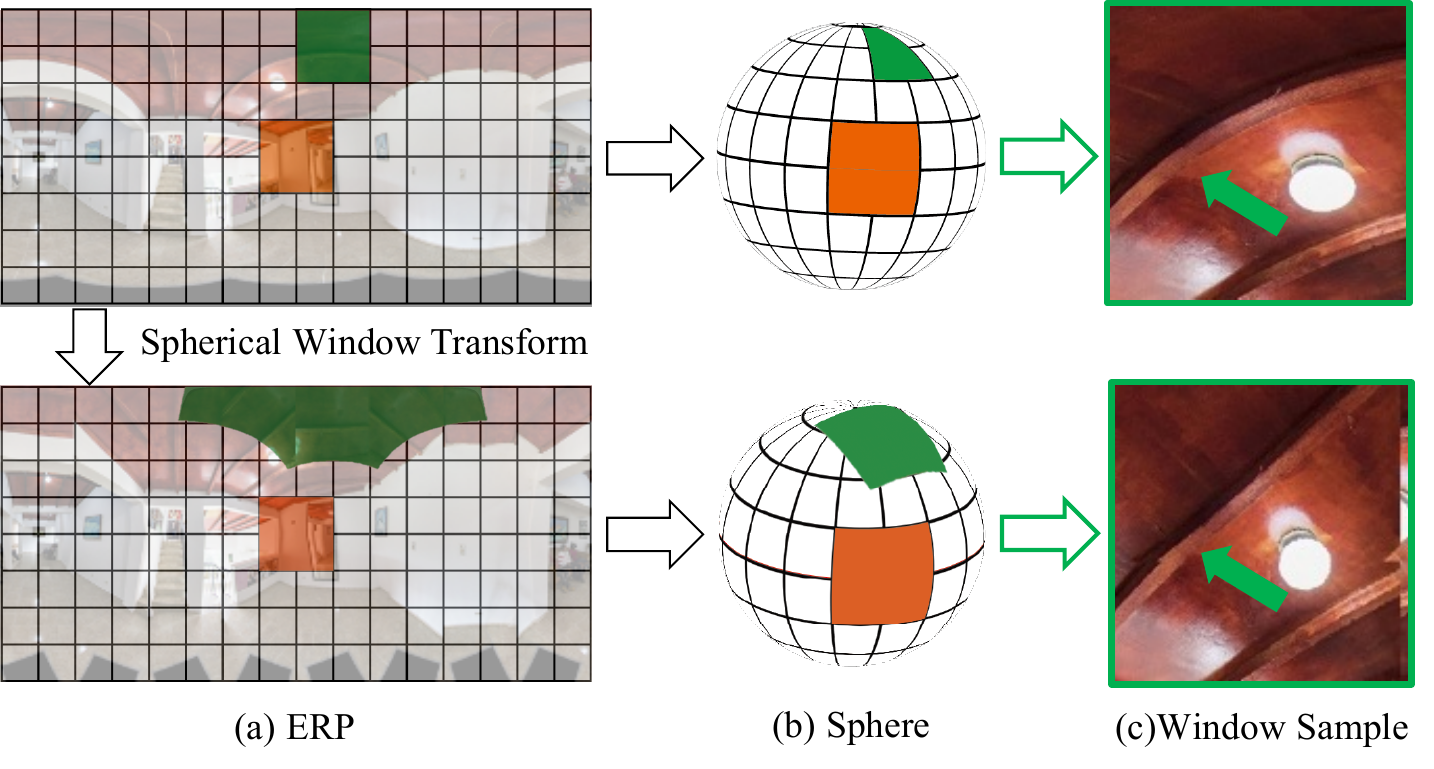}
\vspace{-15pt}
\caption{\textit{\textbf{First row}}: ERP image exhibits severe distortions. The regular window at the pole involves insufficient spherical points. \textit{\textbf{First column}}: With our proposed spherical window transform (SWT), each window is transformed to have sufficient spherical relationships. It is based on the rotational invariance of the sphere. \textit{\textbf{Last column}}: After transformation, the distortion is significantly reduced. The transformed window has better spherical relationships.}
\label{fig:coverfig}
\end{figure}

\IEEEPARstart{R}{ecently}, $360^{\circ}$ images have attracted significant interest due to their ability to capture a complete surrounding environment in a single shot~\cite{ai2022deep}. Depth prediction from a single $360^{\circ}$ image can augment 3D sensing capabilities and support a variety of applications, such as robotic navigation~\cite{dong2022towards, ming2021deep} and autonomous driving~\cite{schon2021mgnet}. Typically, $360^{\circ}$ images are transmitted and stored with equirectangular projection (ERP) format~\cite{yoon2022spheresr}. However, as shown in Fig.~\ref{fig:coverfig}, ERP images exhibit non-uniform pixel distributions across latitudes and suffer from severe distortions in the polar regions. That is, the adjacent spherical points are separated after being projected to the ERP plane. Given that monocular depth estimation is inherently an ill-posed and ambiguous problem~\cite{yuan2022new, bhat2021adabins}, distortions increase the difficulty of $360^{\circ}$ depth estimation.

To mitigate the distortion issue, several works~\cite{coors2018spherenet, Shen2022PanoFormerPT} have proposed to calculate spherical neighbors for each pixel. For instance, SphereNet~\cite{coors2018spherenet} establishes a tangent patch for each pixel and adjusts the sampling positions of the convolutional kernel according to the tangent patch. Based on the pixel-wise transformer, PanoFormer~\cite{Shen2022PanoFormerPT} samples the eight most relevant tokens for each central token from the tangent domain. However, these methods have two main limitations. \textbf{1)} As a tangent patch has only one common point with the spherical surface, the spherical relationships are constructed neighboring the central pixel. The neighboring spherical relationships make it difficult to capture long-range dependencies effectively. \textbf{2)} The geometric projection is conducted pixel by pixel, which is complicated and time-consuming. As a result, these methods record look-up tables in advance. Moreover, once the input size changes, all the pre-recorded look-up tables are not feasible. The calculation process needs to be conducted from scratch.


Fully Connected Conditional Random Fields (FC-CRFs)~\cite{yuan2022new} construct long-range dependencies by connecting any nodes in the graph. The planar relationships between pixels are extended from neighboring regions to distant regions, benefiting the refinement of depth estimation.
Following~\cite{yuan2022new}, we begin by evenly partitioning an ERP image into regular windows in a non-overlapping manner. The equator windows exhibit less distortion and involve sufficient spherical neighbors. In contrast, the windows at the poles have larger distortion and involve insufficient spherical neighbors (See Fig.~\ref{fig:coverfig}(b)), making it difficult to construct effective spherical relationships. 

To address this issue, we propose Spherical Fully Connected CRFs (SF-CRFs) to \textit{capture better spherical relationships}.
Specifically, to involve sufficient spherical neighbours for each window, we first propose a spherical window transform (\textbf{SWT}) module (Sec.~\ref{sec:swt}). We observe that the equator windows have the least distortion. Therefore, the planar neighboring points in the equator window can maintain similar spherical distances (See Fig.~\ref{fig:coverfig}(a-b)). Based on it, we replicate the equator window's spherical relationships to all other windows. Specifically, we first regard the equator window as a template, and sample nodes uniformly in it. 
Secondly, the sampled nodes in the template are transformed according to the spherical rotation matrix. The spherical relationships between the nodes are kept due to the rotational invariance of the sphere. Finally, by taking the positions of the transformed nodes as indexes, we can sample from the input and generate a transformed window, which has the same center and size as the target window. The transformed window succeeds the spherical relationships of the template window and thus involves sufficient spherical neighbors (See Fig.~\ref{fig:coverfig}(c)). Moreover, to make the SWT module efficient, we decompose the rotation matrix into pitch and yaw matrices. Note that the yaw rotation of the sphere is identical to the horizontal rolling of the ERP image. The decomposition enables the processing time for a $512 \times 1024$ ERP image to be only 0.038 seconds on the CPU.

With the SWT module, we can obtain a pair of regular and transformed windows. The regular window focuses on the local planar relationships in the ERP plane, while the transformed one focuses on the global spherical relationships on the spherical surface. Accordingly, we propose a Planar-Spherical Interaction (\textbf{PSI}) module (Sec.~\ref{sec:psi}) to calculate SF-CRFs between the regular and transformed windows. This way, the weighted features preserve the local details and contain global structural information as well.   

By building a decoder based on the SF-CRFs blocks, we propose \textbf{CRF360D} (Sec.~\ref{sec:crf360d}), a novel $360^{\circ}$ depth estimation framework that achieves state-of-the-art performance across diverse datasets. Our CRF360D is compatible with various perspective image-trained backbones (\eg, EfficientNet), serving as the encoder. We conduct experiments on three datasets. The experimental results demonstrate the superiority of our proposed SF-CRFs and CRF360D framework. Our contributions can be summarised as follows: \textbf{(I}) We propose SF-CRFs, which construct better spherical relationships to empower $360^{\circ}$ depth estimation; (\textbf{II}) We propose the SWT module to improve the spherical neighbors in polar regions and calculate the SF-CRFs with a PSI module; \textbf{(III)} We propose CRF360D, which is compatible with flexible backbone encoders and shows significant superiority on three $360^{\circ}$ depth benchmark datasets.

\section{Related Work}

\noindent\textbf{360$^{\circ}$ Depth Estimation.} 
The degree of distortion in ERP images increases from equator regions to polar regions. To alleviate the effect of distortions, existing methods can be classified into two categories: 1)  Rectify ERP images, which are projected to other distortion-less formats~\cite{Wang2020BiFuseM3, Jiang2021UniFuseUF, Li2022OmniFusion3M}; 2) Rectify network designs, such as the shape~\cite{zioulis2018omnidepth} and sampling positions~\cite{coors2018spherenet} of convolution kernels and attentions. For the first category, BiFuse~\cite{Wang2020BiFuseM3} and UniFuse~\cite{Jiang2021UniFuseUF} extract features with ERP and Cubemap formats simultaneously and fuse them. Inspired by the tangent projection~\cite{coors2018spherenet} that enjoys less distortion than Cubemap, OmniFusion~\cite{Li2022OmniFusion3M} and HRDFuse~\cite{Ai2023HRDFuseM3} project the ERP input to a group of tangent patches, which are processed in parallel and finally merged back to the ERP format. However, the overlapping regions in different tangent patches cause severe discrepancies. Recently, S2Net~\cite{li2023mathcal} employs HealPix~\cite{gorski2005healpix} to convert ERP to uniform spherical points.

For the second category, OmniDepth~\cite{zioulis2018omnidepth} proposes row-wise rectangular convolutional kernels, whose widths vary to adapt to different degrees of distortion. SphereNet~\cite{coors2018spherenet} projects the convolutional kernel on the ERP plane to a plane tangent to the sphere, adapting the sampling positions that are invariant to distortion. PanoFormer~\cite{Shen2022PanoFormerPT} employs the tangent plane to adjust the sampling positions of self-attention. However, as the calculation of tangent planes is complicated, these methods need to construct look-up tables in advance. Once the input resolution changes, the look-up tables are not applicable anymore. Recently, EGFormer~\cite{yun2023egformer} rectifies the relative positional embedding and attention scores in self-attention with spherical coordinates and spherical distances.

\noindent\textbf{Vision Transformer (ViT).}
Vision transformer~\cite{dosovitskiy2020image, liu2021swin} has shown superior performance in various tasks, including monocular depth estimation~\cite{yuan2022new}. To balance the computational costs and performance, Swin Transformer proposes to calculate the attention within windows. To adapt to objects with different sizes, Zhang et al.~\cite{zhang2024vision} proposed to learn adaptive window configurations from data. However, it conducts uniform planar sampling, which can not adapt to the uneven distributions in $360^{\circ}$ images. In addition, deformable attention~\cite{xia2022vision} is introduced to learn offsets for each point. However, the learned offsets are restricted to a small margin.

\section{Preliminaries}

We briefly review the classic Conditional Random Fields (CRFs), and their variants, including FC-CRFs and window FC-CRFs. The CRFs are modeled as an undirected graph, where nodes are pixels and edges are the relationships between pixels. By capturing the spatial relationships, CRFs are effective to refine the depth prediction~\cite{yuan2022new}. Given a depth prediction $\mathbf{x}$, the optimization process is accomplished with an energy function $E(\cdot)$:

\begin{equation}
    E(\mathbf{x}) = \sum_i \psi_u(x_i) + \sum_{ij} \psi_p(x_i, x_j),
\end{equation}

\noindent where $x_i$ and $x_j$ are depth values of node $i$ and $j$, respectively. $\psi_u$ is the unary potential function, which optimizes from individual nodes. $\psi_p$ is the partial potential function that takes into account the relationships between nodes. It is common to calculate the relationships between the current node and its neighboring nodes. To enhance the relationships, FC-CRFs connect the current nodes with any other nodes in the graph. Meanwhile, FC-CRFs raise the problem of expensive computational costs.

To leverage the strong representations of FC-CRFs and reduce the computational complexity, NeWCRFs~\cite{yuan2022new} partition the depth feature into multiple windows and perform FC-CRFs in each window. Specifically, given the depth feature $F\in\mathbb{R}^{H\times W\times C}$, it is evenly partitioned with windows in a non-overlapping manner. Supposing that the each window contains $M \times M$ nodes, $F$ is partitioned to $\{F_w^i\in\mathbb{R}^{M\times M\times C}|i\in[1, N]\}$, where $N$ is the total number of windows. Then, NeWCRFs exploit the multi-head attention mechanism~\cite{vaswani2017attention} for capturing better relationships between nodes.

With the usage of window FC-CRFs, the computational costs decrease to linear to the spatial size of the depth map. However, the even arrangement of windows would result in windows at the poles involving insufficient spherical neighbors. In this case, the FC-CRFs in regular windows can not capture enough contextual information to cope with severe distortions in $360^{\circ}$ images.

\section{The Proposed SF-CRFs and CRF360D Framework}

We first introduce the spherical window transform (SWT) module, that improves the spherical neighbours in each window (Sec.~\ref{sec:swt}). Then, we introduce the planar-spherical interaction (PSI) module (Sec.~\ref{sec:psi}), which achieves the spherical fully-connected CRFs (SF-CRFs) between regular and transformed windows. Finally, we describe the details of our CRF360D framework design (Sec.~\ref{sec:crf360d}).

\begin{figure*}[t]
    \centering
\includegraphics[width=.8\textwidth]{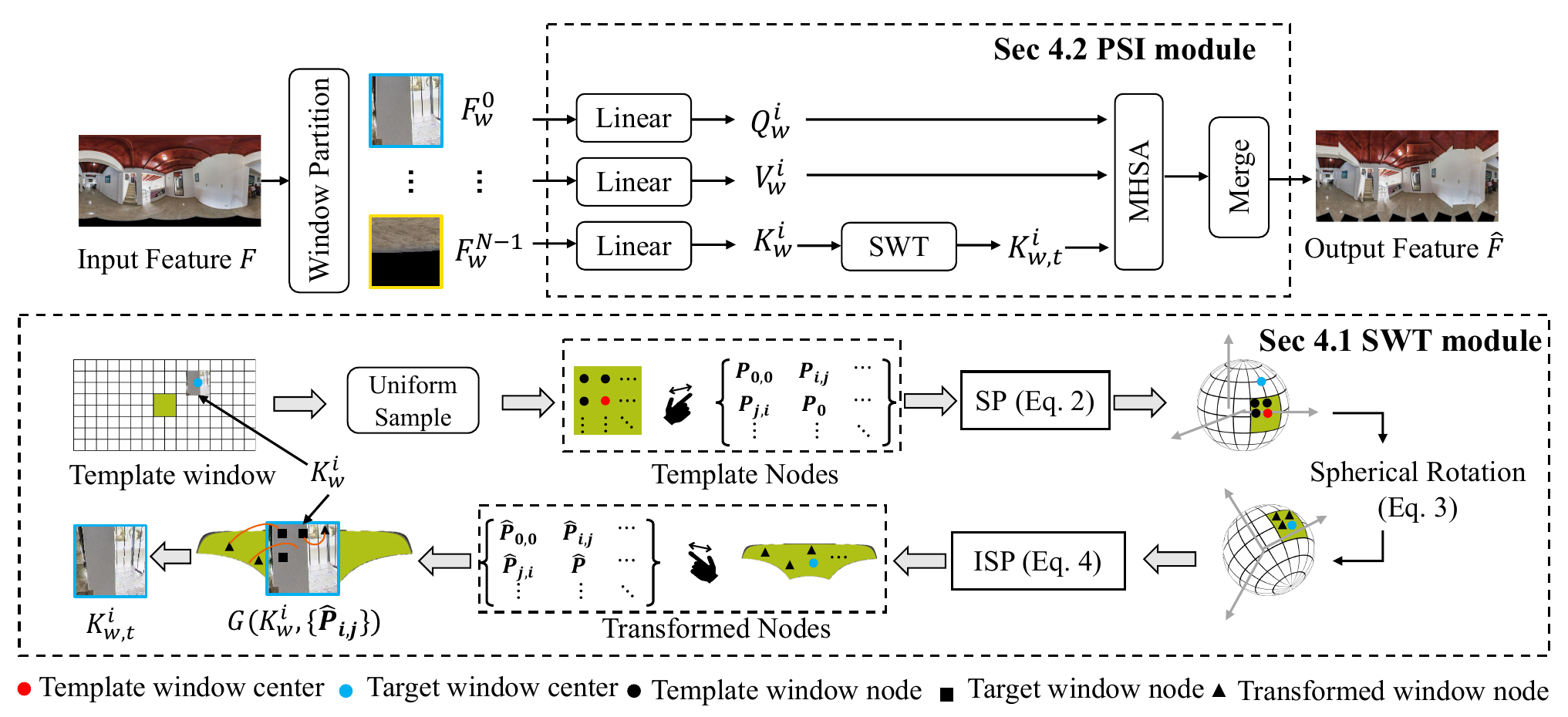}
\vspace{-8pt}
\caption{Illustration of the SWT module and PSI module of our proposed SF-CRFs.}
\label{fig:swt}
\end{figure*}

\subsection{Spherical Window Transform}
\label{sec:swt}

It aims to improve the spherical neighbors in each window. Previous methods~\cite{coors2018spherenet, Shen2022PanoFormerPT} search for spherical neighbors within the tangent domain. However, tangent projection is time-consuming and complicated. Our inspiration is that the equator window exhibits less distortion, whose spherical relationships undergo a small change after being projected from the sphere to the ERP plane, as illustrated in Fig.~\ref{fig:coverfig}. In other words, if we uniformly sample $M \times M$ nodes within an equator window, these nodes keep similar spherical distances on the sphere. Therefore, the window at the equator is superior in its sufficient spherical neighbors. Accordingly, instead of leveraging other distortion-less projections, \eg, tangent projection~\cite{Shen2022PanoFormerPT}, we regard the equator window as a template and replicate its spherical relationships to all other windows.

To establish the template, we first define the angle coordinate system: $(\theta, \phi) \in ([0, \pi], [-\pi, \pi])$, where $\theta$ is the latitude, and $\phi$ is the longitude. Accordingly, the equator center is denoted as $P_0: (\theta_0, \phi_0) = (\frac{\pi}{2}, 0)$. This enables us to create a template window with $P_0$ as the center, which includes $M \times M$ nodes. As shown in Fig.~\ref{fig:swt}, We define nodes in the template window as $\{P_{i,j}: (\theta_{i,j}, \phi_{i,j}), i \in [0, M-1], j \in [0, M-1]\}$. By default, we assume that the template window is square, and the nodes in the window are uniformly distributed. Note that the configurations of the template can be adjusted freely, such as the length, width, and dilation rate. 
Now, we project the node $P_{i,j}$ to the world coordinate system with spherical projection (SP), which is defined as:


\begin{equation}
\text{SP}: \begin{pmatrix}
x_{i,j}  \\
     y_{i,j} \\
     z_{i,j} \\
\end{pmatrix} = 
\begin{pmatrix}
     \cos(\theta_{i,j})\cos(\phi_{i,j}) \\
     \cos(\theta_{i,j})\sin(\phi_{i,j}) \\
     \sin(\theta_{i,j}) \\
\end{pmatrix}.
\label{eq:1}
\end{equation}

\noindent Then, for the target window with center $\hat{P}: (\hat{\theta}, \hat{\phi})$, we aim to transform the template window to its center and search for spherical neighbors for it. Based on the rotational invariance of the sphere, transforming from the equator to the target position only requires a rotation matrix. Specifically, we decompose the rotation matrix with yaw and pitch matrices. The yaw  angle $\alpha$ can be calculated through the difference between $P_0$ and $\hat{P}$: $\alpha = \hat{\phi} - \phi_0$. Similarly, the pitch angle $\beta$ can be calculated through $\beta = \hat{\theta} - \theta_0$. With the rotation angles $\alpha$ and $\beta$, we can conduct the spherical transformation from $(x_{i,j}, y_{i,j}, z_{i,j})$ to $(\hat{x}_{i,j}, \hat{y}_{i,j}, \hat{z}_{i,j})$ with rotation matrix $R$. $R$ is formulated as:

\begin{equation}
R = \begin{pmatrix}
     \cos(\beta) & 0 & -\sin(\beta) \\
     0 & 1 & 0 \\
     \sin(\beta) & 0 & \cos(\beta) \\
\end{pmatrix}
\begin{pmatrix}
     \cos(\alpha) & \sin(\alpha) & 0 \\
     -\sin(\alpha) & \cos(\alpha) & 0 \\
     0 & 0 & 1 \\
\end{pmatrix}
\label{eq:2}
\end{equation}

Then, we project $\hat{P}_{i,j}$ back to the ERP plane via inverse spherical projection (ISP), formulated as follows: 

\begin{equation}
\text{ISP}: \begin{pmatrix}\hat{\theta}_{i,j}\\
\hat{\phi}_{i,j}
\end{pmatrix}=\begin{pmatrix}
\arcsin(\hat{z}_{i,j}) \\
\arctan(\hat{y}_{i,j} /\hat{x}_{i,j}
)\end{pmatrix}.
\end{equation}

\begin{figure*}[t]
    \centering
\includegraphics[width=.8\linewidth]{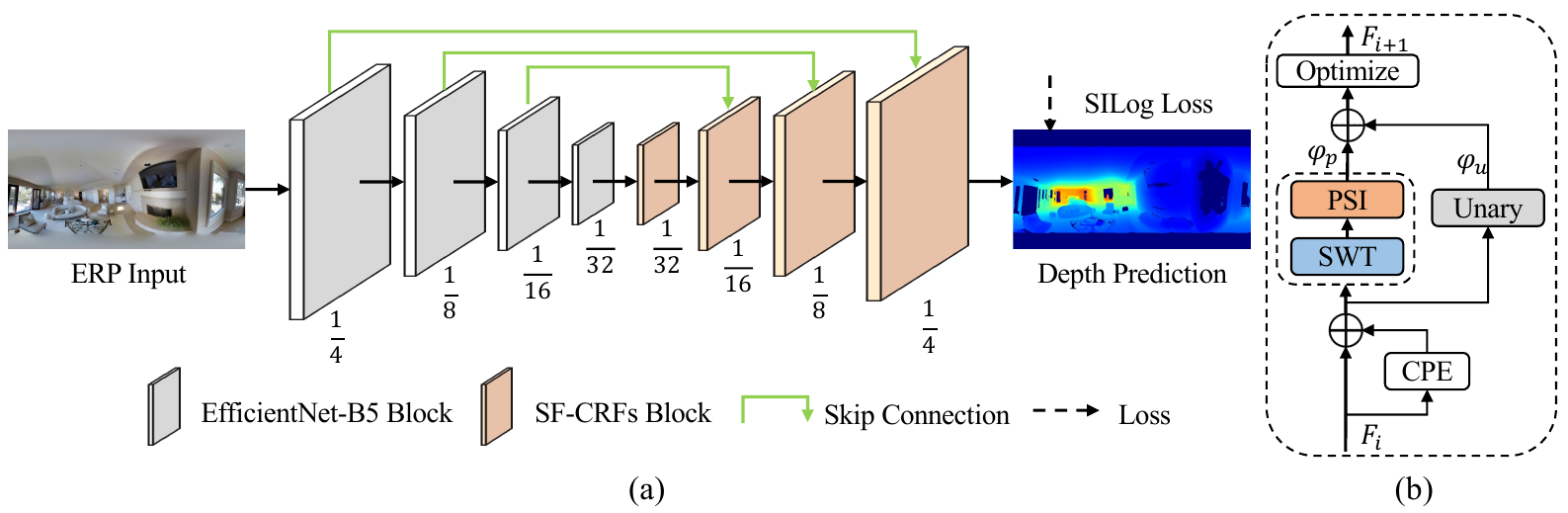}
\vspace{-10pt}
\caption{(a) The overall pipeline of the proposed CRF360D. (b) The architecture of the proposed SF-CRFs.}
\label{fig:framework}
\end{figure*}

After the transformations for all nodes in the template, we obtain  $M \times M$ transformed nodes $\{\hat{P}_{i,j}, i \in [0, M-1], j \in [0, M-1]\}$.
Taking the positions of the transformed nodes as indexes, we can sample from the input feature $F$ with Nearest interpolation. The sampling process can be denoted as $G(F, \hat{P}_{i,j})$. After the sampling, we can obtain a transformed window with center $\hat{P}$ and involves $M \times M$ spherical neighbouring nodes. Furthermore, we notice that the yaw rotation is identical to the horizontal rolling of the ERP image. Therefore, we first transform the windows for $h$ times along the latitude and then roll the $h$ transformed windows for $w$ times. In this case, the transformation times can be decreased from $nH * nW$ to $nH + nW$. As a result, processing a $512 \times 1024$ ERP image only requires 0.038 seconds on the CPU.

\subsection{Planar-Spherical Interaction (PSI) Module}
\label{sec:psi}

After obtaining the regular partitioned and spherical transformed windows, we now establish interactions between these windows to \textit{capture better spherical relationships}. The PSI module plays a role as the partial potential function in the SF-CRFs. As shown in Fig.~\ref{fig:swt}, the input feature $F$ is partitioned into multiple windows $F_w^i$. Then, we calculate the query $Q_w^i$, key $K_w^i$, and value $V_w^i$ through individual projection layers: $Q_w^i=Linear_1(X_w^i); K_w^i=Linear_2(X_w^i); V_w^i=Linear_3(X_w^i)$.
Next, we implement SWT towards $K_w^i$ and \textbf{obtain transformed key feature $K_{w,t}^i$}.
Finally, we utilize the multi-head attention to calculate the weighted window feature:

\begin{equation}
\hat{F}_w^i=MHSA(Q_w^i,K_{w,t}^i,V_w^i).
\end{equation}

After that, we merge window-wise feature $\hat{F}_w^i$ according to spatial relationships and obtain the weighted feature $\hat{F}$. The regular windows contain planar relationships, which can capture local details. In contrast, the transformed windows contain spherical relationships, which can capture global structures. By calculating the correlation between the regular and transformed windows, our proposed PSI module can refine the depth feature with both local details and global structures. By building a decoder based on the SF-CRFs blocks,
we propose CRF360D in the following section.

\subsection{CRF360D Framework}
\label{sec:crf360d}

\noindent \textbf{Overview.} An overview of our proposed CRF360D is depicted in Fig.~\ref{fig:framework}. We adopt an encoder-decoder structural network for monocular $360^{\circ}$ depth estimation. Specifically, for the encoder, we employ EfficientNet-B5 by default. It balances the performance and computational complexity, making our method have comparable computational complexity with existing $360^{\circ}$ monocular depth estimation methods. For the decoder, we stack four levels of SF-CRFs to progressively refine the depth features. Each of the SF-CRFs receives image features from the encoder with a skip connection and the output from the last SF-CRFs block. For an input ERP with spatial resolution $H \times W$, the least spatial resolution for the decoder is $\frac{H}{32} \times \frac{W}{32}$, and the largest spatial resolution is $\frac{H}{4} \times \frac{W}{4}$. For window size $M \times M$, we choose 4 as the default. Note that this choice is smaller than the popular window size, such as 7. We find this choice encourages more transformations and performs better.

\noindent \textbf{Decoder Block.} Each decoder block contains two successive SF-CRFs. The unary potential function is achieved with an identical matrix for efficiency. And the partial potential function is achieved with our SF-CRFs block. The outputs of the two potential functions are added together and fed into an optimization network with a multi-layer perceptron (MLP).

\noindent \textbf{Positional Embedding.} The absolute positional embedding benefits the correlation between the regular and transformed windows. Specifically, we employ conditional positional embedding (CPE)~\cite{chu2021conditional}, which depends on the input depth features and provides additional spatial relationships.

\noindent \textbf{Training Loss.} We use a Scale-Invariant Logarithmic (SILog) loss for supervised training. With predicted depth $D$ and ground-truth depth $\hat{D}$, the logarithm difference is $\begin{aligned}\Delta D_i=\log\hat{D_i}-\log D_i^*\end{aligned}$, where $i$ is the pixel position. We only consider the observed regions in the ground-truth depth map, and the loss is formulated as follows:

\begin{equation}
\mathcal{L}=\alpha\sqrt{\frac1K\sum_i\Delta D_i^2-\frac\lambda{K^2}(\sum_i\Delta D_i)^2},
\end{equation}

\noindent where $K$ is the number of pixels in the observed regions. The hyper-parameters $\alpha$ and $\lambda$ are set to 10 and 0.85, respectively, following the previous work~\cite{yuan2022new}.

\begin{table*}[!t]
    \centering
    \caption{Quantitative comparison with the SOTA methods. $*$ represents re-training with methods with their default settings. We separately compare with methods w/ and w/o median aligning, which will influence the performance. \textcolor{green}{Green} boxes indicate the best performance.
    }
    \vspace{-8pt}
    \label{tab:deph/comparison-depth}
    \resizebox{0.8\textwidth}{!}{ 
    \begin{tabular}{c|c|c|c|c|c|c|c|c|c}
    \toprule
    Datasets&Method & Publish & Align & Abs Rel $\downarrow$& Sq Rel $\downarrow$& RMSE $\downarrow$ &$\delta_1$ $\uparrow$ & $\delta_2$ $\uparrow$ & $\delta_3$ $\uparrow$\\
    \midrule
    \multirow{9}*{Stanford2D3D} 
&BiFuse~\cite{Wang2020BiFuseM3}&CVPR'20&\multirow{4}*{$-$}&0.1209&$-$&0.4142&86.60&95.80&98.60\\
&UniFuse~\cite{Jiang2021UniFuseUF}&RAL'21&&0.1114&$-$&0.3691&87.11&96.64&98.82\\
&HoHoNet~\cite{Sun2020HoHoNet3I}&CVPR'21&&0.1014&$-$&0.3834&90.54&96.93&98.86\\
&CRF360D (Ours) & $-$ & &\cellcolor{green!40}\textbf{0.0888} & 0.0510 & \cellcolor{green!40}\textbf{0.3091} & \cellcolor{green!40}\textbf{91.31} & \cellcolor{green!40}\textbf{97.62} & \cellcolor{green!40}\textbf{99.36} \\
\cmidrule{2-9}
&OmniFusion~\cite{Li2022OmniFusion3M}&CVPR'22&\multirow{4}*{$\checkmark$}&0.0950&0.0491&0.3474&89.88&97.69&99.24\\&PanoFormer~\cite{Shen2022PanoFormerPT}&ECCV'22&& 0.1131&0.0723  & 0.3557& 88.08& 96.23&98.55\\
    &HRDFuse~\cite{Ai2023HRDFuseM3} &CVPR'23&&0.0935&0.0508&0.3106&91.40 & 97.98&99.27\\

    &CRF360D (Ours) & $-$ & &\cellcolor{green!40}\textbf{0.0845} & \cellcolor{green!40}\textbf{0.0431} & \cellcolor{green!40}\textbf{0.2932} & \cellcolor{green!40}\textbf{92.53} & \cellcolor{green!40}\textbf{98.38} & \cellcolor{green!40}\textbf{99.44} \\
    \midrule
    
    \multirow{10}*{Matterport3D}
&BiFuse~\cite{Wang2020BiFuseM3}&CVPR'20&\multirow{6}*{$-$}&0.2048&$-$&0.6259&84.52&93.19&96.32\\
&UniFuse~\cite{Jiang2021UniFuseUF}&RAL'21&&0.1063&$-$&0.4941&88.97&96.23&98.31\\
&HoHoNet~\cite{Sun2020HoHoNet3I}&CVPR'21&&0.1488&$-$&0.5138&87.86&95.19&97.71\\
&NeWCRFs~\cite{yuan2022new} & CVPR'22 & & 0.0906 & $-$ & 0.4778 & 91.97 & \cellcolor{green!40}\textbf{97.61} & 99.09 \\
    &S2Net~\cite{li2023mathcal} & RAL'23& & 0.0911 & $-$ & 0.4280 & 91.90 & $-$ & $-$ \\
& CRF360D (Ours) & $-$ & &\cellcolor{green!40}\textbf{0.0891} & 0.0677 & \cellcolor{green!40}\textbf{0.4241} & \cellcolor{green!40} \textbf{92.16} & 97.47 & \cellcolor{green!40}\textbf{99.17} \\ 
    \cmidrule{2-9}
    &OmniFusion~\cite{Li2022OmniFusion3M} &CVPR'22&\multirow{4}*{$\checkmark$}&0.1007&0.0969&0.4435&91.43&96.66&98.44\\
&PanoFormer~\cite{Shen2022PanoFormerPT}&ECCV'22&& 0.0904&0.0764& 0.4470& 88.16& 96.61& 98.78\\
&HRDFuse~\cite{Ai2023HRDFuseM3} &CVPR'23&&0.0967&0.0936&0.4433&91.62&96.69&98.44\\
& CRF360D (Ours) & $-$ & & \cellcolor{green!40}\textbf{0.0744} & \cellcolor{green!40}\textbf{0.0551} & \cellcolor{green!40}\textbf{0.3836} & \cellcolor{green!40}\textbf{93.63} & \cellcolor{green!40}\textbf{98.34} & \cellcolor{green!40}\textbf{99.45} \\ 
    \midrule
    
    \multirow{4}*{Structured3D}
    &UniFuse*~\cite{Jiang2021UniFuseUF}&RAL'21 & \multirow{4}*{$-$} & 0.0448 & 0.0067 & 0.0555 & 98.05 & 99.43 & 99.72\\
    &PanoFormer*~\cite{Shen2022PanoFormerPT}&ECCV'22& & 0.0940 & 0.0196 & 0.1057 & 92.60 & 98.13 & 99.17 \\
    &EGFormer*~\cite{yun2023egformer} &ICCV'23 & & 0.0906 & 0.0172 & 0.0998 & 92.63 & 98.20 & 99.22\\
    & CRF360D (Ours) & $-$ & & \cellcolor{green!40}\textbf{0.0385} & \cellcolor{green!40}\textbf{0.0048} & \cellcolor{green!40}\textbf{0.0532} & \cellcolor{green!40}\textbf{98.39} & \cellcolor{green!40}\textbf{99.53} & \cellcolor{green!40}\textbf{99.79} \\
    
    \bottomrule
    \end{tabular}}
    \vspace{-10pt}
\end{table*}

\section{Experiments}

\subsection{Implementation Details}

\noindent \textbf{Dataset.} We conduct on three datasets: Stanford2D3D~\cite{armeni2017joint}, Matterport3D~\cite{chang2017matterport3d} and Structured3D~\cite{zheng2020structured3d} datasets. These three datasets are all real-world datasets. The spatial resolution of $360^{\circ}$ images and depth maps is $512 \times 1024$. For Stanford2D3D and Matterport3D datasets, we split the training, validation, and test sets following~\cite{Jiang2021UniFuseUF}. For the Structured3D dataset, we follow the official split~\cite{zheng2020structured3d} for training, validating, and testing.

\noindent \textbf{Implementation.} We conduct experiments on a single NVIDIA 3090 GPU. By default, we utilize EfficientNet-B5~\cite{tan2019efficientnet} pre-trained on the ImageNet dataset as the backbone for the encoder, which balances the performance and efficiency. We use Adam optimizer~\cite{kingma2014adam} with a constant learning rate of 1e-4. The batch size is 4. We train the three datasets for 60 epochs. For data augmentation, we utilize random color adjustment, horizontal translation, and flipping, following~\cite{Jiang2021UniFuseUF}.

\noindent \textbf{Metrics.} Following~\cite{Jiang2021UniFuseUF, Ai2023HRDFuseM3}, we evaluate the performance with standard metrics including Absolute Relative Error (Abs Rel), Squared Relative Error (Sq Rel), Root Mean Squared Error (RMSE), and three percentage metric $\delta_i$, where $i \in \{1.25^1, 1.25^2, 1.25^3\}$. Note that several methods~\cite{Li2022OmniFusion3M, Ai2023HRDFuseM3} utilize Median Align to reduce the scale differences between the predicted depth and ground truth. For a fair comparison, we evaluate methods w/ and w/o the alignment process separately.

\subsection{Quantitative and Qualitative Evaluation}

\begin{figure*}[t]
    \centering
\includegraphics[width=0.83\textwidth]{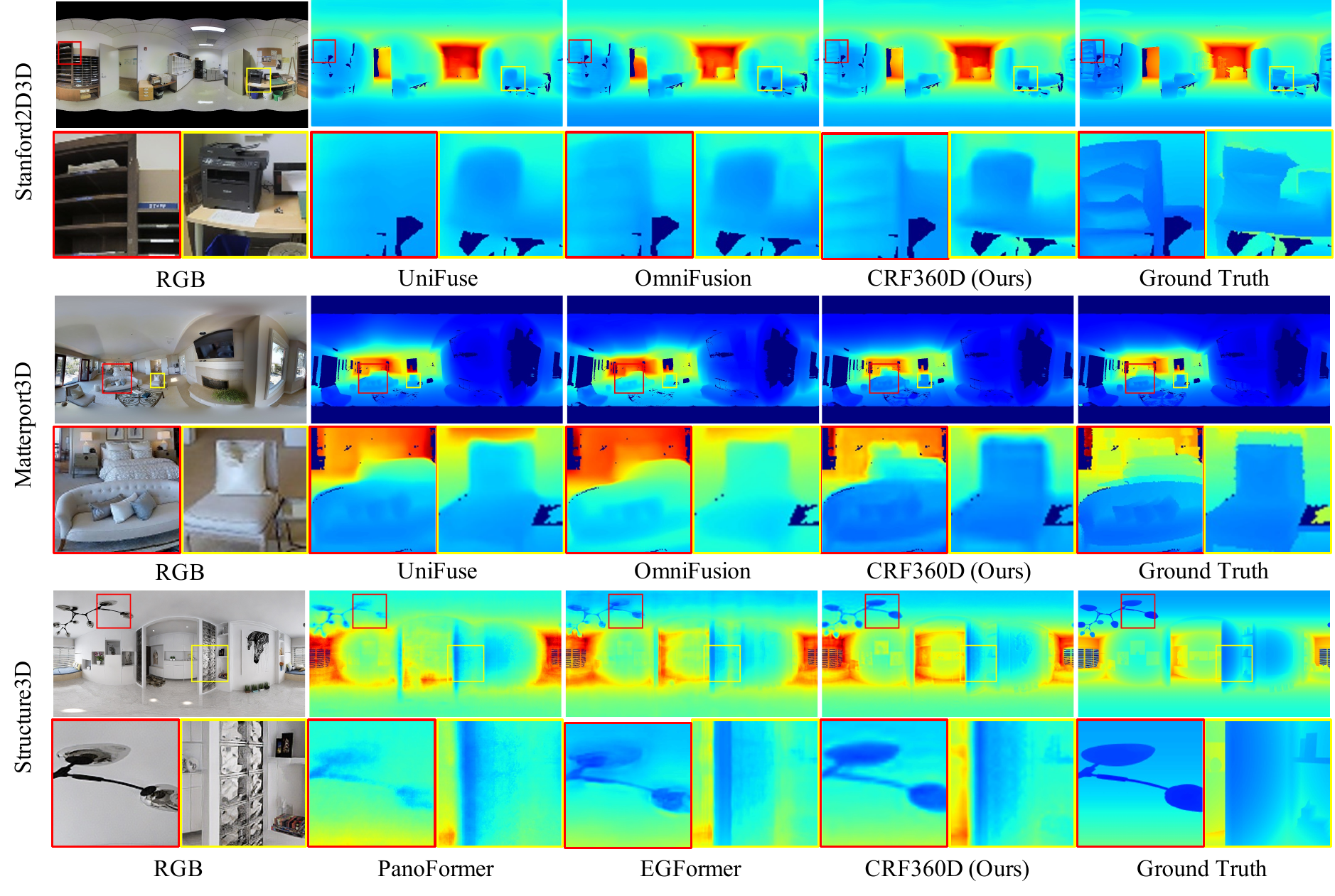}
\vspace{-10pt}
\caption{Qualitative results on Stanford2D3D (top), Matterport3D (middle), and Structured3D (bottom) datasets.}
\label{fig:comparison}
\vspace{-10pt}
\end{figure*}

\textbf{Comparison with SOTA methods.} Tab.~\ref{tab:deph/comparison-depth} provides a quantitative comparison of different methods on three datasets. Our CRF360D outperforms current $360^{\circ}$ depth estimation methods~\cite{Jiang2021UniFuseUF,Wang2020BiFuseM3,Sun2020HoHoNet3I,Li2022OmniFusion3M,Shen2022PanoFormerPT,Ai2023HRDFuseM3,yun2023egformer,li2023mathcal} \textit{in all metrics and datasets}. For example, compared with HRDFuse~\cite{Ai2023HRDFuseM3}, CRF360D has a 9.6\% improvement of Abs Rel metric on the Stanford2D3D dataset. It reveals the effectiveness of our CRF360D capturing effective spherical relationships. Compared with NeWCRFs, CRF360D obtains an 11.2\% gain in the RMSE metric on the Matterport3D dataset. We ascribe that our SF-CRFs focus on capturing spherical relationships, which are superior to NeWCRFs that capture planar relationships. For the Structured3D dataset, we find that PanoFormer and EGFormer show lower performance compared with UniFuse~\cite{Jiang2021UniFuseUF}. We think it is because PanoFormer and EGFormer are based on pure transformer architectures. Without backbones that are pre-trained on perspective images, these methods are difficult to converge to satisfying results. Also, the pixel attention in the two methods is difficult to capture effective contextual information. We also provide qualitative comparisons in Fig.~\ref{fig:comparison}, our CRF360D predicts clearer structural details.

\textbf{Comparison on different backbones.} In Tab.~\ref{tab:ab-backbone} we report our SF-CRFs with various backbones as the encoder, including ResNet-18, ResNet-34, and EfficientNet-B5. We also report the results of baselines, which replace SF-CRFs with feature additions in the decoder. It can be seen that our SF-CRFs show superior results in different backbones. For example, the SF-CRFs obtain a 1.63\% gain in the $\delta_1$ metric. Our CRF360D obtains the best results with EfficientNet-B5 as the encoder.


\subsection{Ablation Studies}

\vspace{3pt}
\noindent \textbf{Spherical Window Transform.} Our transform is based on the regular window partitions. By default, we set the window size as $8 \times 8$, which is similar to the current mainstream choice and is divisible by the size of ERP images. From Tab.~\ref{table:ablation-attention}, we can see that the SWT module obtains a 0.0054 gain in the RMSE metric. In addition, the nodes in the window can be adjusted with learnable offsets~\cite{xia2022vision}. The learnable offsets have performance gains but are inferior to our SWT module. We analyze it because the learnable offsets are limited to a small range, \eg, 2 pixels, which can not address the large distortion at the poles. Moreover, the whole ERP image can be rotated for once with $90^{\circ}$ pitch rotation. Although it can rectify the polar regions, other regions in the ERP image are distorted after the rotation. Therefore, our SWT is superior to one rotation to tackle the distortion in each local window. Note that Tab.~\ref{table:ablation-attention} can also demonstrate the effectiveness of our PSI module, as the regular window with only planar knowledge is inferior to combining both spherical and planar knowledge. 



\vspace{3pt}
\noindent \textbf{Window size.} The common perception is that the performance can be improved with a larger window size, which can directly increase the receptive fields. However, in Tab.~\ref{table:ablation-attention}, we find that $4 \times 4$ window size performs better than $8 \times 8$ window size. We think it is because the smaller window size can encourage more frequent window transformations.

\begin{table}[t]
\setlength{\tabcolsep}{5.5mm}
\centering
\caption{Ablation studies on different window transformations.}
    \vspace{-8pt}
   \label{table:ablation-attention}
   \begin{tabular}{c | c | c}
   \toprule
    Methods & Win. Size & RMSE $\downarrow$\\
    
   \midrule
   Regular window~\cite{liu2021swin} & $8 \times 8$  & 0.4351 \\
   $+$Learnable offsets~\cite{xia2022vision} & $8 \times 8$ & 0.4322 \\
   $+$Pitch rotation~\cite{ling2023panoswin} & $8 \times 8$ & 0.4312\\
       
    $+$SWT (Ours) & $8 \times 8$ & \textbf{0.4297} \\
    \midrule
    Regular window~\cite{liu2021swin} & $4 \times 4$ & 0.4309\\
     $+$SWT (Ours) & $4 \times 4$  & \textbf{0.4241} \\
   \bottomrule
   \end{tabular}
    \vspace{-10pt}
\end{table}

\begin{table}[!t]
\setlength{\tabcolsep}{1mm}
    \centering
    \caption{Comparison of different backbones on Matterport3D test set. 
    }
        \vspace{-8pt}
    \label{tab:ab-backbone}
    \begin{tabular}{c|c|c|c|c|c}
    \toprule
    Backbones & Methods & Param. & FPS & Abs Rel $\downarrow$& RMSE $\downarrow$\\
    \midrule
    \multirow{2}*{ResNet-18~\cite{he2016deep}} & Baseline & \textbf{14.3} & \textbf{41.7} & 0.1276 & 0.5288 \\
    ~ & Ours & 53.2 & 23.1 & \textbf{0.1000} & \textbf{0.4755} \\
    \midrule
     \multirow{2}*{ResNet-34~\cite{he2016deep}} & Baseline & \textbf{24.4} & \textbf{32.8} & 0.1051 & 0.4752 \\
    ~ & Ours & 63.3 & 21.9 & \textbf{0.0961} & \textbf{0.4540} \\
    \midrule
     \multirow{2}*{EfficientNet-B5~\cite{tan2019efficientnet}} & Baseline & \textbf{37.6} & \textbf{22.4} & 0.0964 & 0.4588  \\
    ~ & Ours & 71.3 & 7.7 & \textbf{0.0891} & \textbf{0.4241} \\
    \bottomrule
    \end{tabular}
    \vspace{-8pt}
\end{table}

\noindent \textbf{Different spherical transformations.} The ERP image can be projected to other distortion-less formats, such as cube maps and tangent patches. However, the geometric projection is time-consuming. In Tab.~\ref{table:transform_time}, projecting a $512 \times 1024$ ERP image into six cube map images and re-projecting the cube map images to the ERP plane require 0.25 seconds. Moreover, projecting to 18 tangent patches and merging the patches to the ERP plane require 1.057 seconds in total. In contrast, our SWT module leverages the rotational invariance and transforms the whole ERP image with only 0.038s.


\vspace{-5pt}
\section{Conclusion}

In this paper, we proposed a spherical fully-connected CRFs (SF-CRFs) to capture better spherical relationships. We found that regular windows at the poles were shrunken after being projected to the sphere, causing weak spherical relationships. In contrast, the windows at the equator had the least distortion and presented better spherical relationships. We then replicate the equator's spherical relationships to all other regions. The replication is feasible due to the rotational invariance of the sphere. By building a decoder with SF-CRFs blocks, our method achieved the SOTA results across diverse datasets.

\noindent \textbf{Limitation and Future Work:} Our spherical window transform is accomplished by first establishing a template window at the equator. The template is limited to a single configuration in our method, which is not effective for addressing various objects in the $360^{\circ}$ images. In the future, we will try to combine multiple styles of templates to enrich the representation ability of our SF-CRFs. Furthermore, we will attempt to deploy our method to robotic platforms for real-world applications.

\begin{table}[t]
\setlength{\tabcolsep}{3.5mm}
  \centering
  \caption{Time comparison on different spherical transform methods.}
  \vspace{-8pt}
   \label{table:transform_time}
   \begin{tabular}{c | c | c | c}
   \toprule
   Methods  & Cube Map & Tangent & SWT (Ours) \\
   \midrule
   Time (Seconds) & 0.250 & 1.057 & 0.038 \\
   \bottomrule
   \end{tabular}
    \vspace{-8pt}
\end{table}

%
%
\newpage
\appendix






\begin{abstract}

Due to the lack of space in the main paper, we provide more details of the dataset, metrics, the proposed SWT module, and experimental results in the
supplementary material. Specifically, in Sec.~\ref{sec:dataset}, we provide more details of the dataset and metrics. In Sec.~\ref{sec:swt}, we provide more visualization results of window transformation and the explanation of rotation decomposition. Then, in Sec.~\ref{sec:time}, we provide more comparisons about time consumption between tangent projection and our SWT module. Finally, we provide more quantitative and qualitative results in Sec.~\ref{sec:experiment}.

\end{abstract}

\subsection{Dataset and Metric}
\label{sec:dataset}

\noindent \textbf{Dataset.} We utilize three real-world datasets, \ie, Stanford2D3D~\cite{armeni2017joint}, Matterport3D~\cite{chang2017matterport3d}, and Structure3D~\cite{zheng2020structured3d}, for training, validation, and testing. There are other $360^{\circ}$ depth datasets, such as 3D60~\cite{zioulis2018omnidepth} and PanoSUNCG~\cite{wang2018self} datasets. However, 3D60 has the problem of information leakage due to its rendering, which is not fair for comparison~\cite{Ai2023HRDFuseM3}. In addition, PanoSUNCG is no longer publicly available due to the license issue. Following UniFuse~\cite{Jiang2021UniFuseUF}, we split  Stanford2D3D and Matterport3D datasets. As for the Structure3D dataset, we follow its official splits, where the first 3000 scenes are utilized for training, the middle 250 scenes are utilized for validation, and the last 250 scenes are utilized for testing. We also remove the invalid scenes. The splits of three datasets are listed in Tab.~\ref{table:datasets}.

\begin{table}[ht]
\caption{The splits of the training, validation, and testing sets in three datasets.}
  \centering
  {
  \begin{tabular}{c | c | c | c}
  \hline
 ~ & \multicolumn{3}{c}{Dataset} \\
  \cline{2-4}
  ~ & Stanford2D3D~\cite{armeni2017joint} & Matterport3D~\cite{chang2017matterport3d}& Structure3D~\cite{zheng2020structured3d} \\
  \hline
  Training & 1000 & 7829 & 18298 \\
  \hline
  Validation & 40 & 947 & 1776 \\
  \hline
  Testing & 373 & 2014 & 1691 \\
  \hline
  \end{tabular}}
\label{table:datasets}
\end{table}

\noindent \textbf{Metrics.} We evaluate with standard metrics including Absolute Relative Error (Abs Rel), Squared Relative Error (Sq Rel), Root Mean Squared Error (RMSE), and three percentage metric $\delta_m$, where $m \in \{1.25^1, 1.25^2, 1.25^3\}$. We only calculate the observed pixels in the ground truth depth $D^{*}$. The number of observed pixels is denoted as $K$. Given the predicted depth $D$, the metrics are calculated as follows:

\begin{itemize}
    
    \item Absolute Relative Error (Abs Rel):
    \begin{equation}
    \frac{1}{K}\sum_{i=1}^{K}\frac{||D(i) - D^{*}(i)||}{D^{*}(i)}.
    \end{equation}

    \item Squared Relative Error (Sq Rel):
    \begin{equation}
    \frac{1}{K}\sum_{i=1}^{K}\frac{||D(i) - D^{*}(i)||^2}{D^{*}(i)}.
    \end{equation}
    
    \item Root Mean Square Error:
    \begin{equation}
    \sqrt{\frac{1}{K}\sum_{i=1}^{K}|| D(i) - D^{*}(i)||^2}.
    \end{equation}

    \item $\delta_m$, the fraction of pixels where the relative error between $D$ and $D^{*}$ is less than the threshold $1.25^m$, $m \in \{1,2,3\}$.
    \begin{equation}
    \text{max}\{\frac{D(i)}{D^{*}(i)}, \frac{D^{*}(i)}{D(i)}\} < m.
    \end{equation}
\end{itemize}

As for the median alignment, some $360^{\circ}$ methods~\cite{Li2022OmniFusion3M, Ai2023HRDFuseM3} utilize it to decrease the scale difference between depth prediction and ground truth. The median alignment can influence the final performance, and we separately report results w/ and w/o median alignment. The median alignment is achieved by adjusting the prediction $D$ as follows:

\begin{equation}
    D = D * \frac{\text{median}(D^{*})}{\text{median}(D)}.
\end{equation}

\begin{figure}[t]
    \centering
\includegraphics[width=\linewidth]{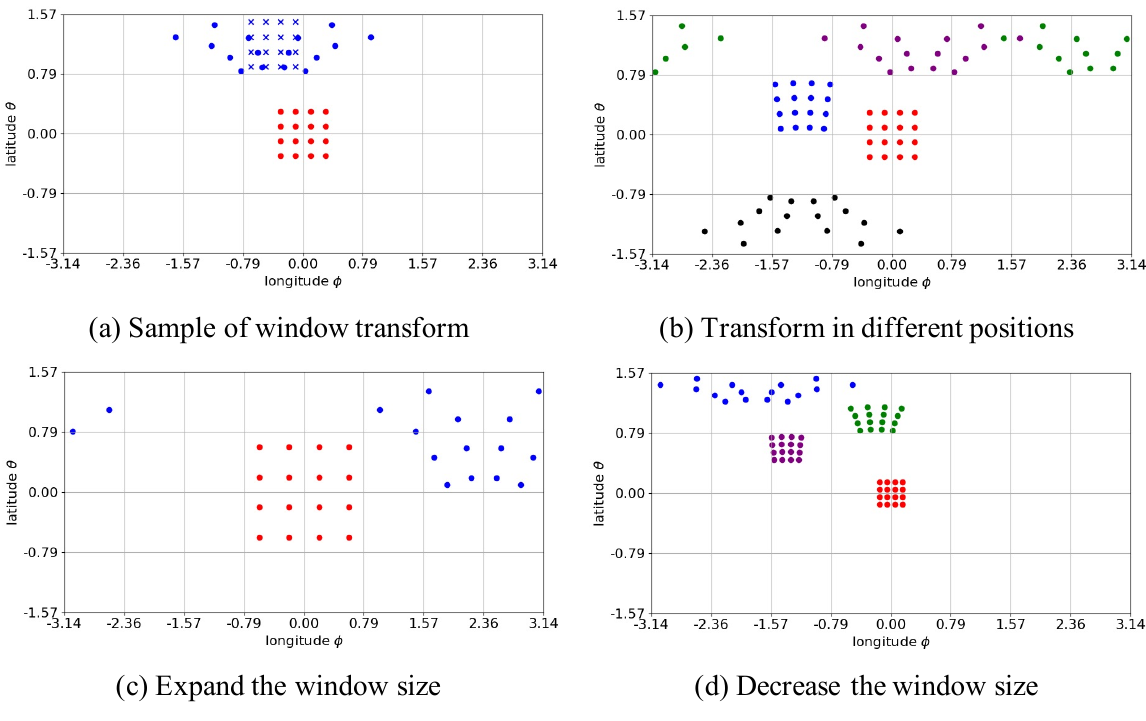}

\caption{Visualization of the window transformation results in the SWT module. Red dots  \textcolor{red}{\scalebox{0.85}[1]{$\bullet$}}: Nodes in the template window; Blue crosses  \textcolor{blue}{\scalebox{0.85}[1]{$\times$}}: Nodes in the target window; Dots with other colors \textcolor{blue}{\scalebox{0.85}[1]{$\bullet$}} \textcolor{Fuchsia}{\scalebox{0.85}[1]{$\bullet$}} \textcolor{ForestGreen}{\scalebox{0.85}[1]{$\bullet$}} \textcolor{black}{\scalebox{0.85}[1]{$\bullet$}}: Nodes in the transformed windows.}
\label{fig:window}
\end{figure}

\subsection{Spherical Window Transform (SWT) module}
\label{sec:swt}
We give more transformation examples in the supplementary material. As shown in Fig.~\ref{fig:window}(a), in the high-latitude region, the transformed nodes are scattered according to the distortion. In this case, the transformed window can involve more spherical neighbours and capture better spherical relationships. In contrast, the regular window can capture local details and planar relationships. 

As shown in Fig.~\ref{fig:window}(b), we show the transformed windows in different positions. Firstly, we can find that from the equator to the poles, the planar distance between transformed nodes increases gradually. Secondly, in the same latitude, different transformed windows can be converted with \textit{horizontal rolling}. Nodes in these transformed windows have consistent spherical relationships. In addition, the discontinuity of left and right boundaries in the ERP image can be eliminated naturally. Thirdly, the transformed windows are longitudinally symmetrical.

As shown in Fig.~\ref{fig:window}(c)(d), we show the transformation results with expanded and decreased window sizes, respectively. Our SWT module can transform windows with different sizes to obtain better spherical relationships. The choice of window size is a balance between receptive field and transformation frequency. With the window size increasing, the receptive field increases, and distant nodes can participate in the interaction. However, the transformation frequency decreases, which makes our SWT module less effective. In the experiment, we find that window size with $4 \times 4$ obtains the best performance.

\begin{figure}[ht]
    \centering
\includegraphics[width=\linewidth]{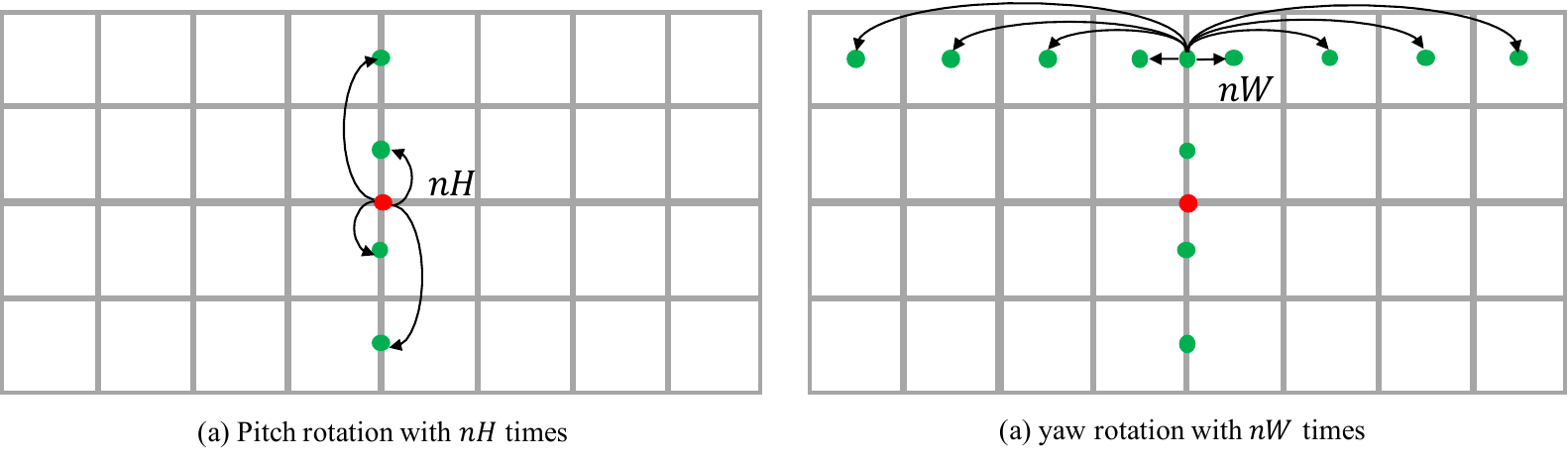}

\caption{Illustration of the decomposition for pitch and yaw rotations.}
\label{fig:decompose}
\end{figure}

Based on the horizontal rolling for transformed windows in the same latitude, we can decompose SWT into pitch and yaw rotations. We partition an ERP image with $nH \times nW$ windows. As shown in Fig.~\ref{fig:decompose}, we first transform the template window to target windows in one longitude. In this way, we can obtain $nH$ transformed windows. Then, we roll the $nH$ transformed windows horizontally and obtain the whole $nH \times nW$ windows. The decomposition makes the hand-crafted implementation codes efficient because the recurrent times decrease from $nH * nW$ to $nH + nW$. We attach the related code in the supplementary material.

\begin{figure}[t]
    \centering
\includegraphics[width=\linewidth]{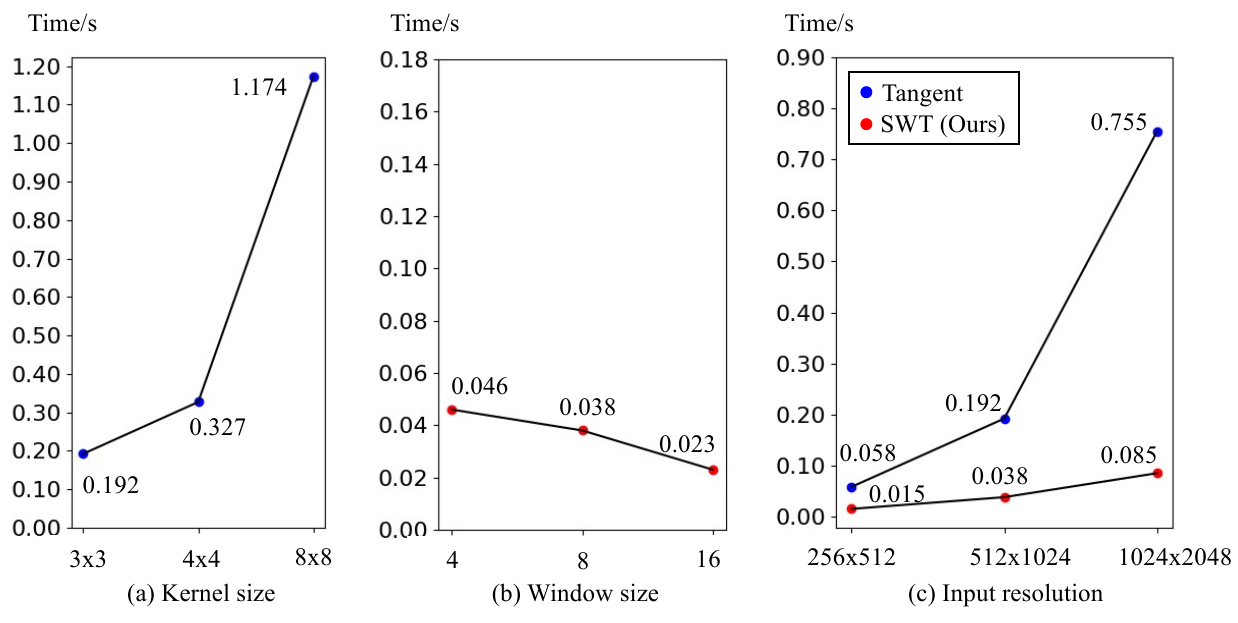}

\caption{Time comparison. (a) Tangent projection with different kernel sizes. (b) Our SWT module with different window sizes. (c) Tangent projection and our SWT module in different input resolutions.}
\label{fig:time}
\end{figure}

\subsection{Time comparison with Tangent Projection}
\label{sec:time}
In Tab.~\textcolor{red}{7} of the main paper, we report the time of projecting an ERP to multiple tangent patches or cube maps. These methods project an ERP into multiple perspective patches. In the supplementary, we provide more comparisons about tangent kernels in PanoFormer~\cite{Shen2022PanoFormerPT} and our SWT. The tangent kernel~\cite{Shen2022PanoFormerPT} is pixel-wise, searching for spherical neighbours for each pixel. In Fig.~\ref{fig:time}(a), we can find that with the kernel size increasing, the time consumption increases rapidly. In contrast, in Fig.~\ref{fig:time}(b), as the window size increases, the time consumption of our SWT module decreases, as a larger window size requires fewer transformation times. Overall, our SWT is more efficient than tangent projection because we focus on window-wise processing instead of pixel-wise processing. Finally, we compare the tangent projection with our SWT module in different input resolutions. The time consumption that our SWT processes a $1024 \times 2048$ ERP is similar to the time consumption that tangent projection processes a $256 \times 512$ ERP image.


\subsection{More Comparison Results}
\label{sec:experiment}
\noindent \textbf{Data Source Explanation.} The Tab.~\textcolor{red}{1} of the main paper reports the performances of different methods on three benchmark datasets. For Stanford2D3D and Matterport3D, most numbers are excerpted from~\cite{Ai2023HRDFuseM3}. Note that~\cite{Ai2023HRDFuseM3} re-trains OmniFusion and PanoFormer. In addition, we report the results of perspective method NeWCRFs~\cite{yuan2022new} on Matterport3D from its paper. For S2Net~\cite{li2023mathcal}, it mainly utilizes Swin Transformer as the backbone, whose parameters and computational complexity are too large compared with other $360^{\circ}$ depth estimation methods. For a fair comparison, we report its results with EfficientNet-B5 as the backbone (See Tab.~\textcolor{red}{VII} from its paper). For Structure3D dataset, we re-train three methods, \ie, UniFuse~\cite{Jiang2021UniFuseUF}, PanoFormer~\cite{Shen2022PanoFormerPT}, and EGFormer~\cite{yun2023egformer}. The main reason is that EGFormer utilizes additional data for training. For a fair comparison, we only utilize the training set of Structure3D for training. The three methods are re-trained based on their official settings.

\begin{table}[t]
\centering
\setlength{\tabcolsep}{5mm}
  \caption{Comparison of different methods about FPS and GFLOPs.}
   \label{table:FPS}
   \begin{tabular}{c || c c}

   \toprule
   Methods  & FPS & GFLOPs \\
   \midrule
    
   BiFuse~\cite{Wang2020BiFuseM3} & 0.9 & 199.6 \\
    HRDFuse~\cite{Ai2023HRDFuseM3} & 5.5 & 50.6 \\
    PanoFormer~\cite{Shen2022PanoFormerPT} & 9.2 & 77.7 \\
    EGFormer~\cite{yun2023egformer} & 9.6 & 73.9 \\
    Ours-ResNet & 23.1 & 108.4 \\
    Ours-EfficientNet & 7.7 & 112.5 \\
   \bottomrule
   \end{tabular}
\end{table}

\begin{table}[t]
\centering
\setlength{\tabcolsep}{10mm}
\caption{Ablation studies on absolute positional embedding.}
     \label{table:ablation-ape}
\begin{tabular}{c || c}

   \hline
   APE methods & RMSE \\
   \hline
   None  & 0.4341 \\
   uvxyz~\cite{ling2023panoswin} & 0.4300 \\
   CPE~\cite{chu2021conditional} & \textbf{0.4241} \\
   \hline
   \end{tabular}
\end{table}

\begin{table}[t]
\centering
\setlength{\tabcolsep}{10mm}
 \caption{Ablation studies on the effect of different loss functions.}
   \label{table:ablation-loss}
\begin{tabular}{c || c}

   \hline
   Loss  & RMSE \\
   \hline
    BerHu~\cite{laina2016deeper}  & 0.4322 \\
   RMSE log~\cite{li2023mathcal} & 0.4387 \\
    SILog~\cite{eigen2014depth} &\textbf{0.4241} \\
   \hline
   \end{tabular}
\end{table}

\noindent \textbf{Running speed and computational complexity.} In Tab.~\ref{table:FPS}, we compare with other methods in running speed and computational costs. It can be seen that our method runs faster than some fusion-based methods, such as HRDFuse~\cite{Ai2023HRDFuseM3} and BiFuse~\cite{Wang2020BiFuseM3}. With ResNet-18 as the encoder, our CRF360D runs faster than PanoFormer~\cite{Shen2022PanoFormerPT} and EGFormer~\cite{yun2023egformer}. Note that the running speed of these two methods is calculated with look-up tables. Moreover, our CRF360D can achieve the best performance across these $360^{\circ}$ depth estimation methods with about an extra 20 GFLOPs.

\noindent \textbf{Absolute positional embedding.} As our PSI module calculates relationships between the regular and transformed windows, learning the absolute positional embedding benefits to decrease the window difference. From Tab.~\ref{table:ablation-ape}, adding absolute planar and spherical coordinates obtains a 0.0041 gain in the RMSE metric. Moreover, by learning the positional embedding according to the input feature~\cite{chu2021conditional}, the performance improves by 0.0059 in the RMSE metric. 

\noindent \textbf{Loss function.} In Tab.~\ref{table:ablation-loss}, the SILog loss has the best performance compared with BerHu and RMSE log losses. the SILog loss is commonly utilized in perspective depth estimation methods~\cite{yuan2022new}, which is also effective for our CRF360D.

\begin{figure*}[t]
    \centering
\includegraphics[width=\textwidth]{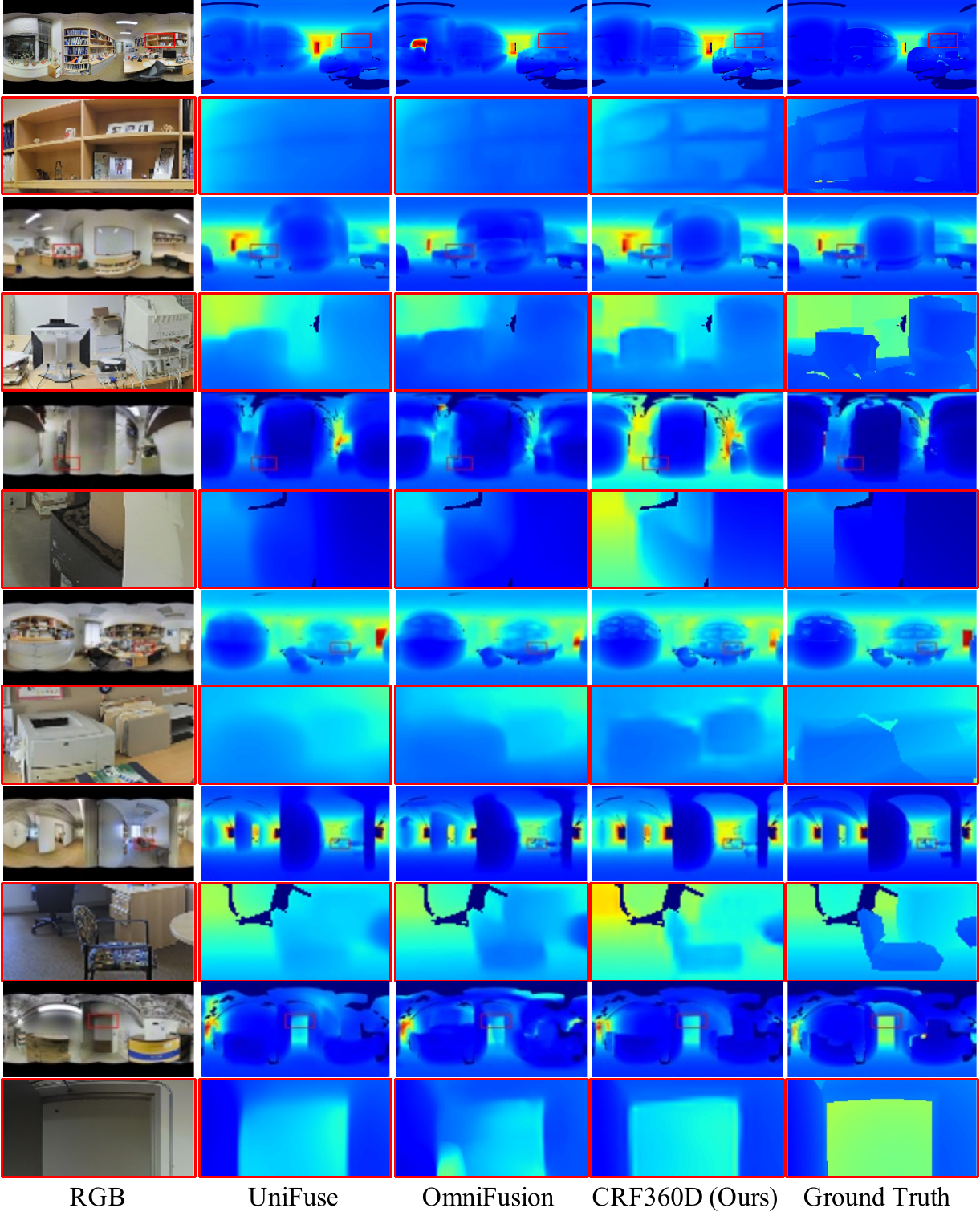}

\caption{Qualitative results on Stanford2D3D dataset.}
\label{fig:supp-stanford2d3d}
\end{figure*}

\begin{figure*}[t]
    \centering
\includegraphics[width=\textwidth]{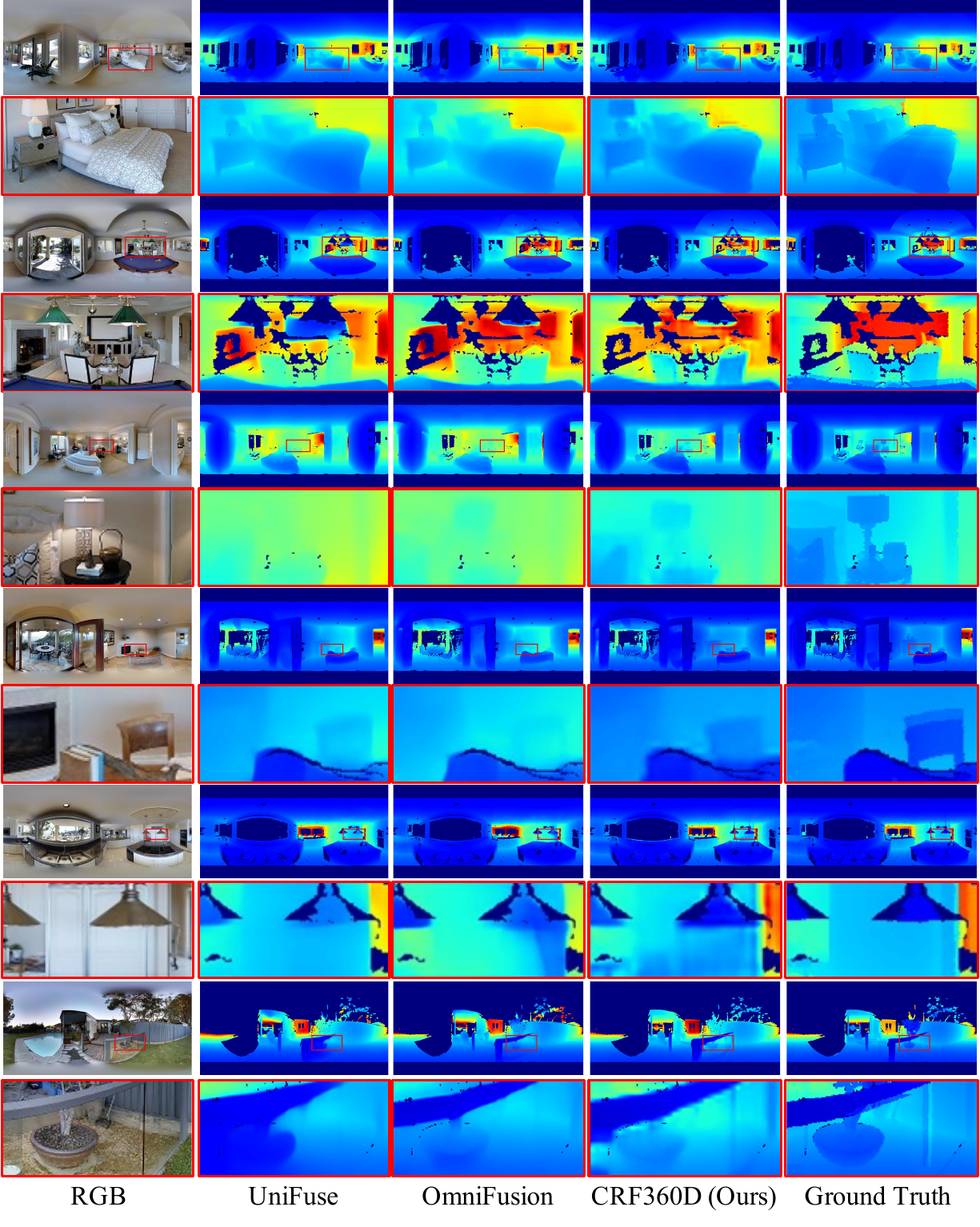}

\caption{Qualitative results on Matterport3D dataset.}
\label{fig:supp-matterport3d}
\end{figure*}

\begin{figure*}[t]
    \centering
\includegraphics[width=\textwidth]{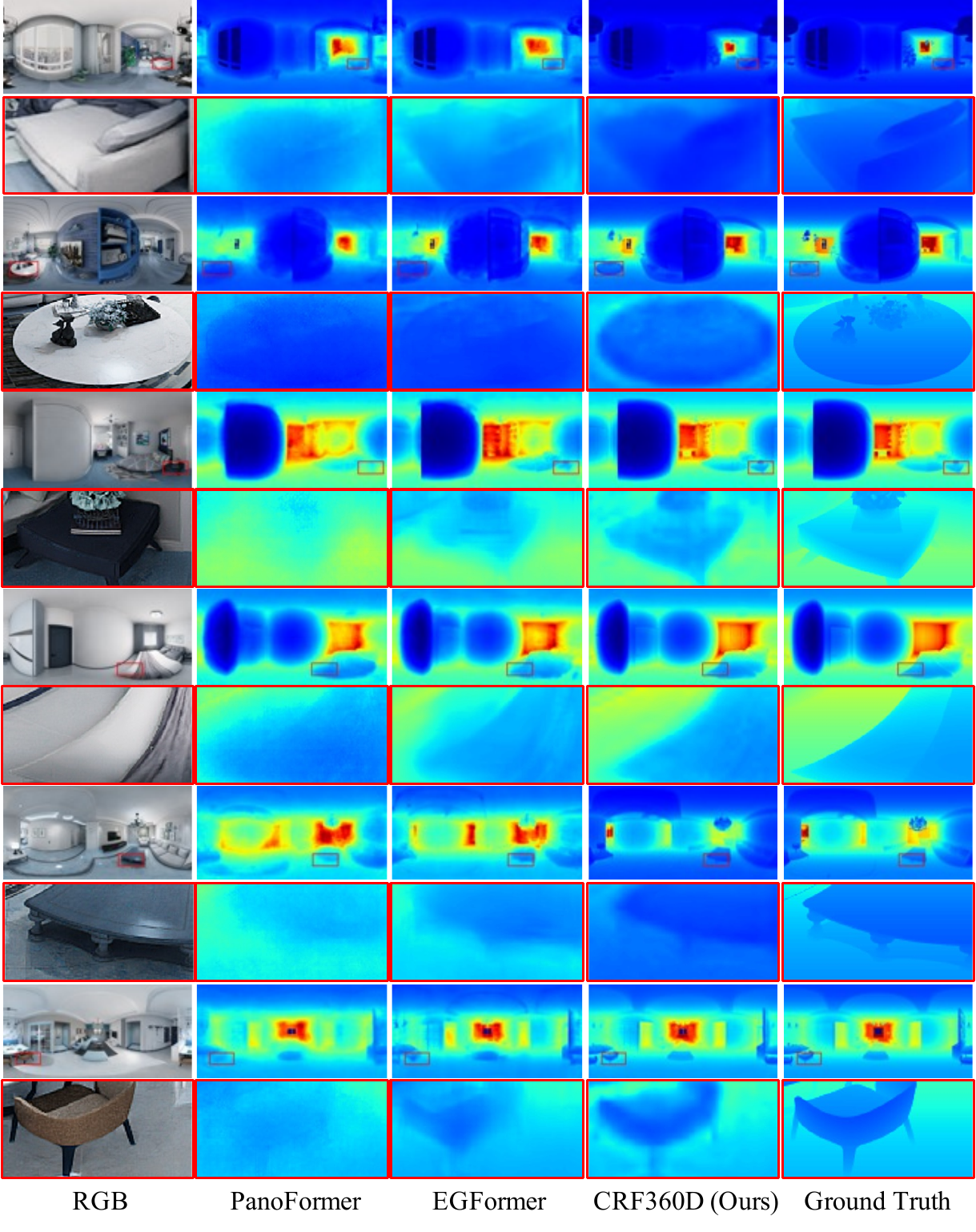}

\caption{Qualitative results on Structure3D dataset.}
\label{fig:supp-structure3d}
\end{figure*}

\noindent \textbf{More visualization results.} In Fig.~\ref{fig:supp-stanford2d3d},~\ref{fig:supp-matterport3d},~\ref{fig:supp-structure3d}, we give more qualitative results on three benchmark datasets. The visualization results demonstrate that our CRF360D predicts clearer structural details than other methods, especially in some small objects, which are often blurry or even invisible in previous methods. For example, in the 8th row of Fig.~\ref{fig:supp-stanford2d3d}, our CRF360D can predict the structural details of the printer and box on the table.

\bibliographystyle{IEEEtran}
\bibliography{IEEEabrv, egbib}
\vfill

\end{document}